\documentclass{article}
\PassOptionsToPackage{numbers, compress}{natbib}
     \usepackage[preprint]{neurips_2019}
\usepackage[utf8]{inputenc} 
\usepackage[T1]{fontenc}    
\usepackage{hyperref}       
\usepackage{url}            
\usepackage{booktabs}       
\usepackage{amsfonts}       
\usepackage{nicefrac}       
\usepackage{microtype}      
\usepackage{amsmath,amssymb,subcaption,float}
\usepackage{graphicx}
\DeclareMathOperator*{\argmax}{\arg\!\max}
\DeclareMathOperator*{\argmin}{\arg\!\min}
\title{Memory-Efficient Episodic Control Reinforcement Learning with Dynamic Online k-means}

\author{%
  Andrea Agostinelli\\
  Department of Bioengineering\\
  Imperial College London\\
  \texttt{aa7918@ic.ac.uk} \\
  \And
  Kai Arulkumaran\\
  Department of Bioengineering\\
  Imperial College London\\
  \texttt{kailash.arulkumaran13@imperial.ac.uk} \\
  \And
  Marta Sarrico\\
  Department of Bioengineering\\
  Imperial College London\\
  \texttt{mvs918@ic.ac.uk} \\
  \And
  Pierre Richemond\\
  Data Science Institute\\
  Imperial College London\\
  \texttt{phr17@ic.ac.uk} \\
  \And
  Anil A. Bharath\\
  Department of Bioengineering\\
  Imperial College London\\
  \texttt{a.bharath@imperial.ac.uk}
}

\begin{document}

\maketitle

\begin{abstract}
Recently, neuro-inspired episodic control (EC) methods have been developed to overcome the data-inefficiency of standard deep reinforcement learning approaches. Using non-/semi-parametric models to estimate the value function, they learn rapidly, retrieving cached values from similar past states. In realistic scenarios, with limited resources and noisy data, maintaining meaningful representations in memory is essential to speed up the learning and avoid catastrophic forgetting. Unfortunately, EC methods have a large space and time complexity. We investigate different solutions to these problems based on prioritising and ranking stored states, as well as online clustering techniques. We also propose a new dynamic online $k$-means algorithm that is both computationally-efficient and yields significantly better performance at smaller memory sizes; we validate this approach on classic reinforcement learning environments and Atari games.
\end{abstract}

\section{Introduction}
Reinforcement learning (RL), using neural networks as function approximators, have surpassed human performance in a wide range of environments \cite{mnih2015human}. However, these approaches are sample-inefficient: they can require hundreds of times the experience of a human to reach the same level of performance during the early stages of learning \cite{lake2017building}. Recently, new neuro-inspired algorithms known as episodic control (EC) methods \cite{blundell2016model,pritzel2017neural}, implementing non-/semi-parametric models, have outperformed the speed of learning of state-of-the-art deep RL algorithms. EC methods rely on looking up past transitions from memory, where novel states are evaluated based on their similarity to past states. The lookups use $k$-nearest neighbours ($k$-NN) search, which is demanding in terms of both space and time complexity, and as the lookup operation is performed every time the agent encounters a new state, this makes it very difficult to scale up existing EC algorithms. 

The novel problem that we aim to address, with the online nature of the data distribution in RL, is how to reduce the size of the memory in EC methods. And while reducing the memory size, the memory structure still has to support the RL agent in learning from recently-observed states without catastrophically losing past knowledge, referred to as the stability-plasticity dilemma in memory retention \cite{abraham2005memory}. In consideration of these problems, our aim is to investigate and design novel memory storage approaches for EC algorithms that retain performance while reducing the amount of memory needed. We make the following contributions: Firstly, we evaluate 5 different memory storage strategies, including a novel online clustering algorithm, applied to 2 EC algorithms, in both classic RL environments and Atari games. Secondly, we show that replacing least-recently-used states or using online clustering techniques achieves the best performance across a range of settings and environments. Finally, we propose a new online clustering algorithm, which outperforms the other memory storage strategies when using smaller memory sizes.

\section{Background}

\textbf{Reinforcement Learning:} RL is the study of optimising the behaviour of an agent embodied in an environment. In the RL framework, the agent observes at timestep $t$ the current \textit{state} $\mathbf{s}_t$, interacts with the environment using \textit{action} $\mathbf{a}_t$, thereby generating the transition to the successive \textit{state} $\mathbf{s}_{t+1}$ and receiving a feedback signal (\textit{reward}) $r_{t+1}$. The behaviour of an agent is controlled by the \textit{policy} $\pi(\mathbf{a}_t|\mathbf{s}_t)$. The final goal is to learn the optimal policy $\pi^*$ that maximises the expected return in the environment; the optimal policy $\pi^*$ is defined as: ${{\pi^*}=\argmax_{\pi}\mathbb{E}[R\mid{\pi}]},$ where the return $R$ is the cumulative, $\lambda$-discounted reward of its sequence of experiences: ${R = \sum_{t=0}^{T-1}{\lambda}^t r_{t+1};\,\lambda \in [0,1]}$.

\textbf{Episodic Control:} Memory and learning are supported in the brain by two main systems, the hippocampus and the neocortex. In particular, neocortical changes are related to long-term memory and learning of statistical models of sensory experiences, while hippocampal activity seems to perform fast instance-based learning on recent experiences; the latter system is thought to be the main location of rapid learning in humans \cite{mcclelland1995there}. Recently the behaviour of the hippocampus has been translated into a novel EC RL algorithm, where Q-value estimation is performed as weighted $k$-NN regression, emulating the hippocampal instance-based retrieval of the past.

\begin{figure*}
\includegraphics[scale=0.26]{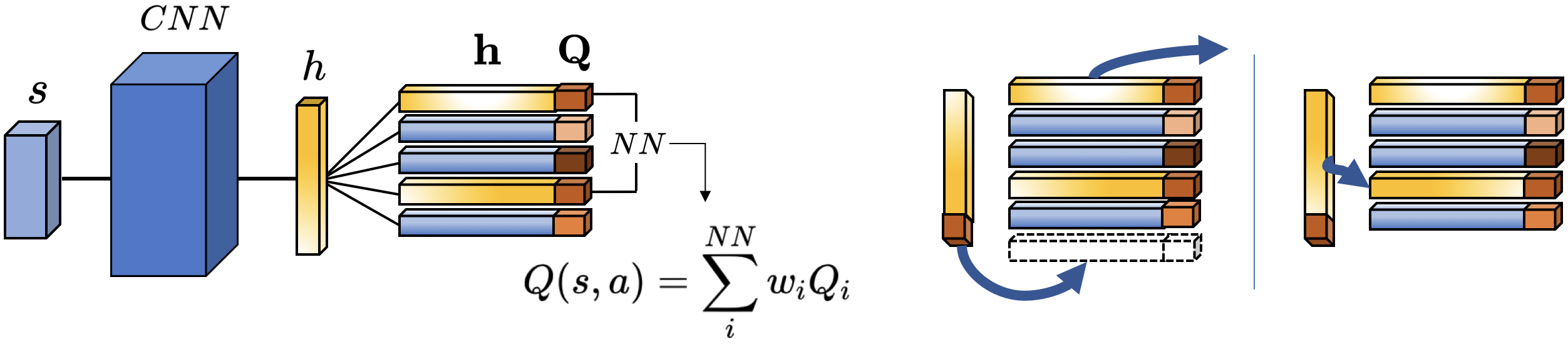}
\centering
\caption{NEC inference and updating. (left) A state $\mathbf{s}$ is turned into an embedding/key $\mathbf{h}$ via a neural network, after which its $Q$-value is estimated through a weighted average among its nearest neighbours (NN). (right) New states are stored either by completely replacing another stored state, or merging it into an existing cluster of states.}
\label{fig:NEC}
\end{figure*}

In model-free EC (\textit{MFEC}) \cite{blundell2016model} the agent is comprised of per-action tables \textbf{$Q^{EC}$} containing a list of the highest returns ever obtained by taking action $\mathbf{a}$ from state $\mathbf{s}$: $Q^{EC}(\mathbf{s}, \mathbf{a})$. MFEC is non-parametric, and uses either Gaussian random projections \cite{johnson1984extensions} or variational autoencoders \cite{kingma2013auto,rezende2014stochastic} to reduce the dimensionality of the observations. A semi-parametric EC model was introduced as neural EC (NEC) \cite{pritzel2017neural}, a deep RL agent that uses a combination of a gradient descent to slowly improve the representation of the state through a neural network, and a quickly-updated value function through ``differentiable neural dictionaries'', similar to MFEC per-action tables. The complete state-value estimation process is shown in Figure \ref{fig:NEC}, where the weights for the average in the $Q$-value calculation are determined by the inverse distance weighted kernel: $w_i = k\left(\mathbf{h}, \mathbf{h}_{i}\right)=\frac{1}{\left\|\mathbf{h}-\mathbf{h}_{i}\right\|_{2}^{2}+\delta}$, where $\mathbf{h}_i$ is a learned embedding of a state $\mathbf{s}_i$. The parameters of NEC are updated through minimising the residual between the current Q-value and $n$-step episodic returns.

\textbf{Memory Storage Strategies:} After filling a memory with finite storage, there are two main strategies to deal with new observations, as illustrated in Figure \ref{fig:NEC}; \textbf{prioritising} the storage of states with specific characteristics and reducing the number of elements with \textbf{clustering techniques}.

In MFEC and NEC the update of the memory is done by removing the least-recently-used entries. Outside of EC, Isele \textit{et al.} \cite{isele2018selective} illustrated four selection strategies to manage the storage of past transitions $\{\mathbf{s}_t,\mathbf{a}_t,\mathbf{s}_{t+1},r_{t+1}\}$ in experience replay \cite{lin1992self}: prioritising surprising states, favouring higher rewards, narrowing the set of experiences trying to match the global distribution of states, and maintaining a heterogeneous distribution of the state-space. 

Generally, when the agent operates in a streaming setting, using data that is not independent and identically distributed (i.i.d.), conventional first-in-first-out (FIFO) memory buffers will fail due to the non-i.i.d. data stream \cite{hayes2019memory}. Clustering algorithms are an effective class of methods that can mitigate catastrophic forgetting, reducing the number of redundant datapoints. Classic clustering techniques have the disadvantage of being applicable only to static sets of data, but online clustering methods can be applied successfully in streaming settings \cite{hayes2019memory}. 

\section{Implemented Strategies}
\subsection{Prioritised Memories}
Using MFEC and NEC as baseline EC algorithms, we investigate the influence of different memory storage strategies. The following strategies apply when the memory is completely filled; a ranking rule prioritises the replacement of the states in the memory.

\textbf{Maximise Surprise (SUR)} \cite{isele2018selective}: States with the highest prediction error (surprise) could be more useful to store, as compared to states that are well-predicted with the current memory. The surprise of a given state is calculated as the absolute value of the difference between the future discounted return and its predicted Q-value; therefore we drop the state $\mathbf{s}$ with the minimum surprise:
$$\argmin_{\mathbf{s} \in \text{memory}}|R - Q(\mathbf{s}, \mathbf{a})|.$$

\textbf{Maximise Rewards (REW)} \cite{isele2018selective}: States with the highest return may be the best to retain. With this strategy the state with the lowest return is replaced:
$$\argmin_{\mathbf{s} \in \text{memory}}R.$$


\textbf{Least Recent Used (LRU)} \cite{blundell2016model}: The state that has been least recently used by the memory is replaced by the new state. This is the default strategy for MFEC and NEC, inspired by the behaviour of the brain in forgetting old experiences.

\subsection{Clustering Techniques}
An alternative to ranking strategies is to reduce the number of states by merging them into clusters, with newly observed states merged with their nearest (Euclidean) neighbour. 

\textbf{Online $k$-means (kM)}: There are several versions of the ``online $k$-means`` algorithms \cite{zhong2005efficient,liberty2016algorithm,king2012online}, but we use the computationally-efficient and simple version from \cite{hayes2019memory}. 
After filling the memory, a separate vector $\textbf{n}$ is used to store the number of elements of each cluster, and after observing a new state the memory is updated by:
$$\mathbf{c}_{i} \leftarrow \frac{n_{i} \mathbf{c}_{i}+\mathbf{s}}{n_{i}+1},\qquad
Q_{i} \leftarrow \frac{n_{i} Q_{i}+R}{n_{i}+1},\qquad
n_{i} \leftarrow n_{i}+1,$$
where $\mathbf{c}_{i}$ is the closest neighbour to the new state $\mathbf{s}$ with associated $Q$-value $Q_i$, and $n_{i}$ is the number of elements of $\mathbf{c}_{i}$, subsequently incremented by one.

\textbf{Dynamic online $k$-means (DkM)}: This is our novel clustering algorithm, based on online $k$-means. We use a new heuristic for the update, based on the memory size $N$:
$$\mathbf{c}_{i} \leftarrow \frac{n_{i} \mathbf{c}_{i}+\mathbf{s}}{n_{i}+1},\qquad
Q_{i} \leftarrow \frac{n_{i} Q_{i}+R}{n_{i}+1},\qquad
n_{i} \leftarrow n_{i}+1,\qquad
\mathbf{n} \leftarrow \mathbf{n}-\frac{1}{N}.$$
In comparison to online $k$-means, the ``size'' of the clusters is reduced over time. Once $n_i \leq 0$ the cluster is completely replaced by the new state $\mathbf{s}$, with $n_{i}$ reset to 1. Our heuristic, inspired by LRU, aims to delete clusters that are not frequently encountered, and also prevents clusters from becoming static.

Once the nearest neighbour is found for a memory of size $N$, kM has an additional complexity of $O(1)$, and DkM has an additional complexity of $O(N)$, which is negligible compared to the original lookup ($O(N^2)$ for naive $k$-means or $O(dNlog(N))$ for $k$-d/cover trees \cite{beygelzimer2006cover}, with $d$ proportional to the state dimensionality).

\section{Experiments}
We investigated the importance of using alternative memory strategies in both simple and complex RL domains, showing results for different memory capacities. We use the original configurations of MFEC and NEC as EC baselines, with the dueling double DQN (D3QN) as a baseline \cite{mnih2015human,hasselt2010double,wang2016dueling}. We tested our implementations in three sets of environments: classic control \cite{brockman2016openai}, room domains (re-implemented from \cite{machado2017laplacian}), and Atari games \cite{bellemare2013arcade}.
For every environment we use a discount factor $\lambda$ of 0.99; a full set of hyperparameters per environment are available in Section \ref{sec:hyperparameters}. Every agent is evaluated with an $\argmax$ policy over 10 different episodes every evaluation interval.

\begin{figure}
  \centering
  \begin{minipage}{0.48\textwidth}
    \includegraphics[width=0.8\textwidth]{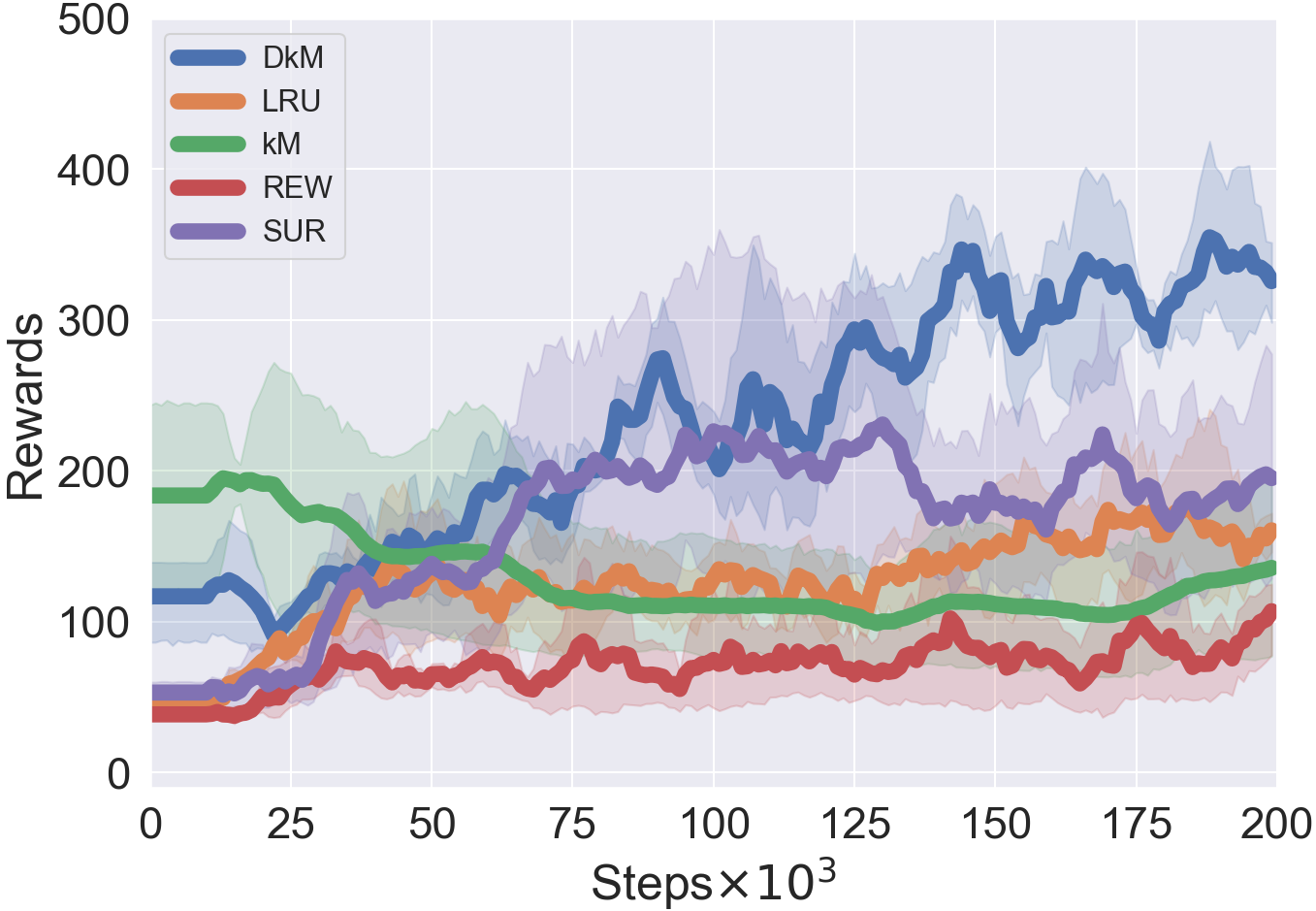}
      \centering
    \caption{Learning on Cartpole using NEC and a small memory size of 50. DkM is the only method that reaches average total rewards > 200.}
    \label{fig:cart50}
  \end{minipage}
  \hfill
  \begin{minipage}{0.48\textwidth}
    \includegraphics[width=0.8\textwidth]{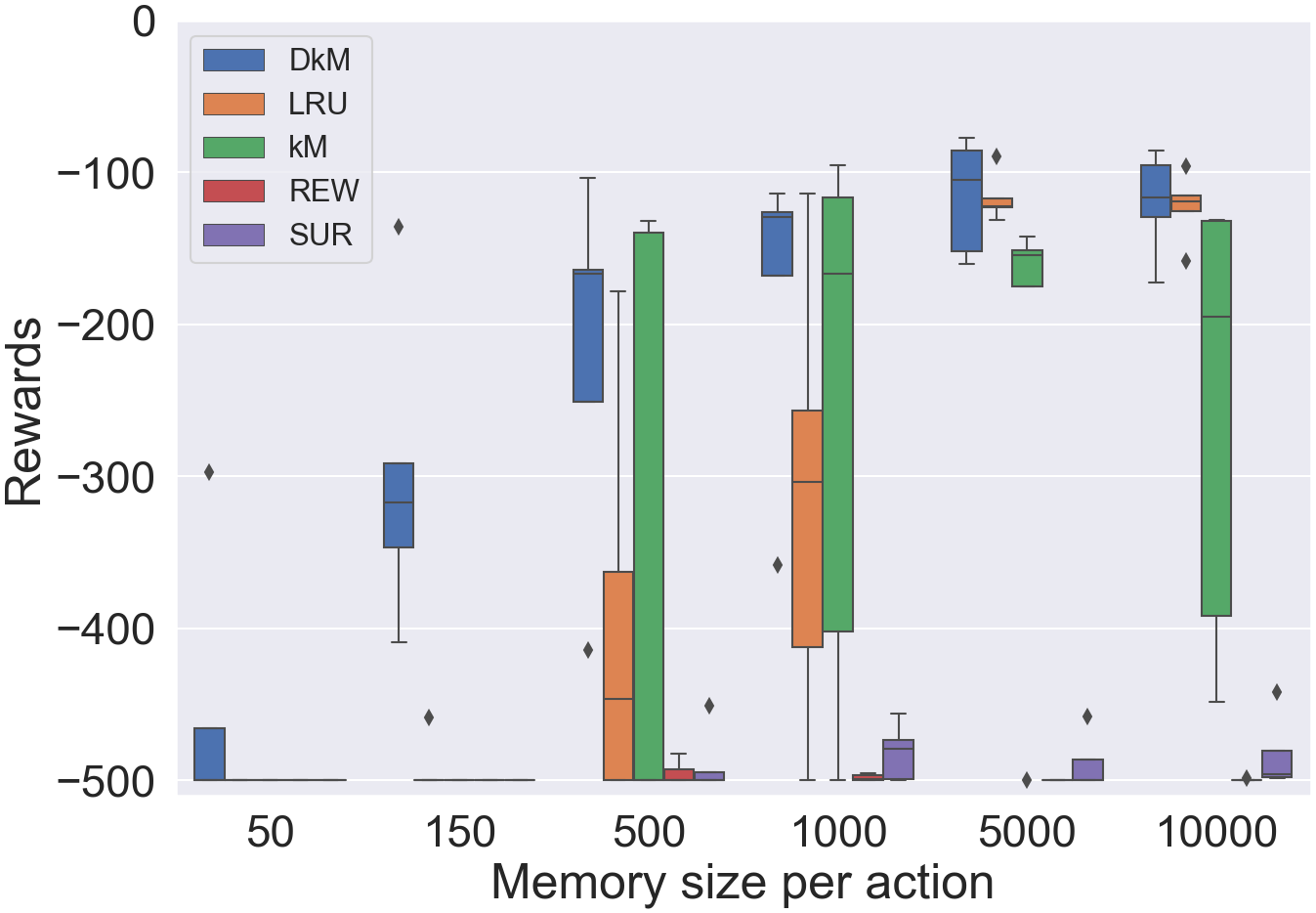}
      \centering
    \caption{Total rewards on Acrobot, using NEC and different memory sizes. Overall, DkM yields high return episodes with low variance.}
    \label{fig:acro_box}
  \end{minipage}
\end{figure}

\textbf{Classic Control}: We considered Acrobot and Cartpole, which have a continuous state space and 3 and 2 actions respectively. Table \ref{table1} summarises results for Cartpole. Clustering techniques perform the best. In particular DkM applied to NEC consistently shows the best performance for low buffer sizes. Memory size 50, where DkM significantly outperforms the other techniques, is illustrated in Figure \ref{fig:cart50}. In Acrobot, DkM outperforms all the other methods, as shown in Table \ref{table2}. Here our novel method achieves the best rewards in 10 out of 12 settings, and always when using memory sizes less than 5000. In the final 2 settings (MFEC with memory sizes 5000 and 10000), DkM is second only to LRU. With NEC, DkM is always the best memory storage method (Figure \ref{fig:acro_box}).

\begin{table*}
\centering
\caption{Total rewards in the Cartpole domain, after $2\times10^4$ (NEC) and $1.5\times10^4$ (MFEC) steps; the MFEC agents typically converged within $0.5\times10^4$ steps, while the NEC agents continued to improve. The values indicate the mean of the last 10 evaluations, averaged over 5 random seeds.}\smallskip
\resizebox{1\textwidth}{!}{ 
\begin{tabular}{ c c c c c c | c c c c c | c}
\hline
  \textbf{Memory Size} & \multicolumn{5}{c}{\textbf{MFEC}} & \multicolumn{5}{c}{\textbf{NEC}}& \multicolumn{1}{c}{\textbf{D3QN}}\\
 \textbf{per Action} &  LRU  & REW & SUR  & kM & DkM &  LRU  & REW & SUR  & kM & DkM\\
 \hline
 \hline
 50 & $96 \pm{21}$& $\pmb{180 \pm47}$& $51 \pm{24}$& $133 \pm{66}$& $123 \pm{13}$& $161 \pm{15}$& $107 \pm{29}$& $195 \pm{88}$& $136 \pm{72}$ & $\pmb{326 \pm31}$\\
 
 100 & $110 \pm{9}$& $134 \pm{25}$& $57 \pm{15}$ & $108 \pm{37}$& $\pmb{153\pm25}$& $235 \pm{53}$& $132 \pm{42}$& $215 \pm{97}$& $216 \pm{159}$& $\pmb{339\pm32}$\\
 
 150 & $117 \pm{13}$& $125 \pm{30}$& $83 \pm{15}$& $74 \pm{21}$& $\pmb{142 \pm26}$ & $246 \pm{22}$& $167 \pm{51}$& $215 \pm{36}$&$151 \pm{56}$& $\pmb{290\pm24}$\\
 
 500 & $142 \pm{19}$& $61 \pm{7}$& $102 \pm{12}$& $\pmb{244\pm131}$& $174 \pm{34}$& $247 \pm{43}$& $117 \pm{33}$& $218 \pm{48}$ & $211 \pm{79}$& $\pmb{271 \pm37}$&  $249$\\
 
 1000 & $\pmb{197\pm37}$& $49 \pm{13}$& $151 \pm{36}$& $196 \pm{26}$& $181 \pm{18}$& $220 \pm{46}$& $131 \pm{39}$& $145 \pm{44}$& $237 \pm{65}$& $\pmb{257 \pm46}$ & $ \pm{143}$\\
 
 3000 & $276 \pm{31}$& $45 \pm{8}$& $183 \pm{18}$& $\pmb{359 \pm81}$& $209 \pm{13}$& $254 \pm{42}$& $110 \pm{44}$& $220 \pm{15}$& $\pmb{300\pm61}$& $255 \pm{50}$\\
 
 5000 & $259 \pm{43}$& $54 \pm{13}$& $210 \pm{23}$& $\pmb{352 \pm86}$& $213 \pm{34}$& $282 \pm{38}$& $114 \pm{14}$& $219 \pm{32}$& $\pmb{343\pm97}$& $302 \pm{53}$\\
 
 10000 & $325 \pm{55}$& $65 \pm{11}$& $266 \pm{28}$& $\pmb{344\pm30}$& $228 \pm{15}$& $309 \pm{23}$& $124 \pm{42}$& $250 \pm{27}$& $\pmb{330 \pm27}$& $265 \pm{25}$\\
 \hline
\end{tabular}
}
\label{table1}
\end{table*}

\begin{table*}
\centering
\caption{Total rewards in the Acrobot domain, after $2\times10^4$ steps. The values indicate the mean of the last 10 evaluations, averaged over 5 initial random seeds for all methods.}\smallskip
\resizebox{1\textwidth}{!}{ 
\begin{tabular}{ c c c c c c | c c c c c |c}
\hline
  \textbf{Memory Size} & \multicolumn{5}{c}{\textbf{MFEC}} & \multicolumn{5}{c}{\textbf{NEC}}& \multicolumn{1}{c}{\textbf{D3QN}}\\
 \textbf{per Action} &  LRU  & REW & SUR  & kM & DkM &  LRU  & REW & SUR  & kM & DkM\\
 \hline
 \hline
 50 & $-500$& $-500$& $-304 \pm{28}$& $-500$& $\pmb{-298\pm108}$& $-500$& $-500$& $-500$& $-500$& $\pmb{-453\pm79}$\\
 
 150 & $-499 \pm{2}$& $-500$& $-431 \pm{136}$& $-500$& $\pmb{-165\pm45}$& $-492 \pm{17}$ & $-500$& $-500$& $-500$& $\pmb{-300\pm91}$\\
 
 500 & $-486 \pm{19}$& $-500$& $-493 \pm{9}$& $-425 \pm{149}$& $\pmb{-164\pm41}$& $-397 \pm{121}$& $-495 \pm{7}$& $-489 \pm{19}$ & $-354 \pm{179}$& $\pmb{-220\pm108}$& $-336$\\
 
 1000 & $-387 \pm{87}$& $-500$& $-479 \pm{42}$& $-430 \pm{141}$& $\pmb{-345\pm190}$& $-317 \pm{132}$& $-498 \pm{2}$& $-482 \pm{17}$ & $-256 \pm{164}$& $\pmb{-179\pm92}$&$\pm{149}$\\
 
 5000 & $\pmb{-368\pm167}$& $-500$& $-499 \pm{1}$& $-478 \pm{44}$& $-397 \pm{155}$& $-116.4 \pm{14}$& $-500$& $-489 \pm{16}$& $-225 \pm{138}$& $\pmb{-115.8\pm34}$\\
 
 10000 & $\pmb{-284\pm173}$& $-500$& $-497 \pm{5}$& $-497 \pm{4}$& $-395 \pm{132}$& $-123 \pm{20}$& $-500$& $-483 \pm{22}$& $-260 \pm{134}$& $\pmb{-120\pm30}$\\
 \hline
\end{tabular}
}
\label{table2}
\end{table*}

\textbf{Gridworld}:  We recreated two room environments, OpenRoom and FourRoom. The agent can move in 4 directions and is given a reward of 1 only when the agent moves to the goal. OpenRoom is a 10x10 open room, and FourRoom is a set of four interconnected rooms, totalling 11x11. Clustering techniques are particularly efficient in the OpenRoom domain, with kM performing the best at memory sizes 2 and 5, and DkM performing best overall across all settings. DkM performs well with a memory size of 10 (or greater), which is 5 times smaller than what is required for the LRU method. However, in FourRoom only LRU was capable of perfectly solving the environment using high memory sizes, bigger than 150 per action. Tables and learning curves are provided in Section \ref{sec:grid}.

\textbf{Atari}: A set of five games with varied gameplay types are evaluated, namely: Ms. Pac-Man, Q*bert, Pong, Space Invaders and Bowling. As in prior work, we use standard preprocessing of the visual observations \cite{mnih2015human}, and use Gaussian random projections for MFEC \cite{blundell2016model}. As preferring either highly rewarding (REW) or highly surprising states (SUR) as the memory storage strategy achieved poor performance in both Classic Control and Gridworld domains, we did not evaluate these methods on Atari games due to limited computational resources. For this reason we also did not evaluate kM, as it was generally outperformed by DkM. We ran experiments for 5 million steps, where every agent step represents 4 game frames (action repeat of 4). Due to limited computational resources we ran experiments with NEC for 3.5 million steps, when using $10^5$ items per action. As in the original papers \cite{blundell2016model,pritzel2017neural}, we use 11 nearest-neighbours and a final $\epsilon$ of 0.005 for MFEC, and 50 nearest neighbours and a final $\epsilon = 0.001$ for NEC. We did, however, use a key size of 128 and the inverse distance weighted kernel for MFEC, as this performed the same or better than a simple average in our initial experiments, and makes these hyperparameters the same as for NEC.\footnote{As a result we do not include the original results as hyperparameter changes mean they are not directly comparable.}

\begin{figure}
  \centering
    \includegraphics[width=\textwidth]{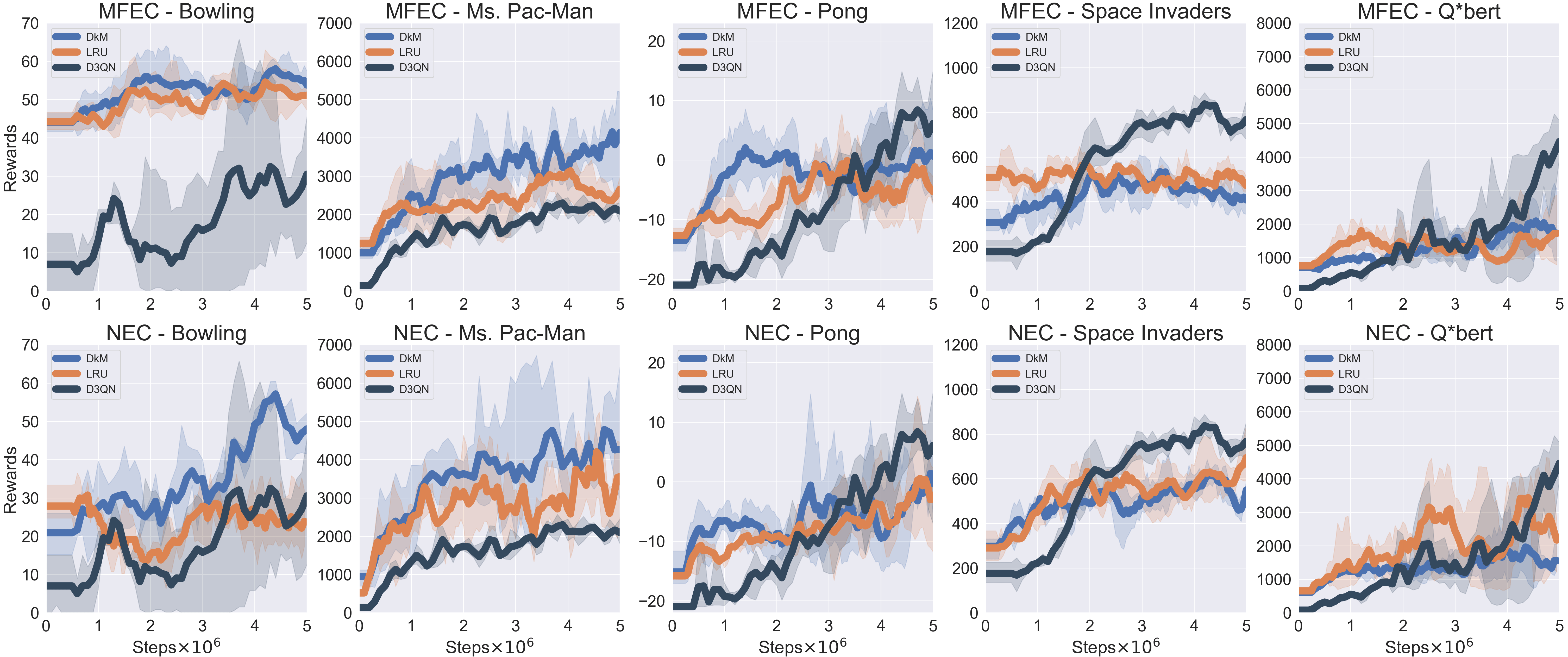}
    \caption{Learning curves of 5 Atari games using MFEC, NEC and D3QN. EC methods with a memory size of $10^4$ per action. DkM generally improves upon LRU.}
    \label{atari10}
\end{figure}

Figure \ref{atari10} shows the learning curve of 5 different Atari games, using a memory size of $10^4$ items per action. DkM performed the best across the experiments, compared to LRU (Table \ref{atari_table}). In Bowling. Ms. Pac-Man and Pong, DkM achieves the best results, for either MFEC and NEC. The performance of EC methods increases using a bigger memory size of $10^5$ items per action, where they show much better sample efficiency in comparison to fully parametric approaches such as the D3QN (Table \ref{atari_table}). EC methods are faster in achieving high rewards. LRU outperforms DkM for high memory size; it always achieves better rewards in Space Invaders and Q*Bert for both MFEC and NEC.

\begin{table}
\centering
\caption{Mean and standard deviations of final scores calculated across 3 seeds on 5 Atari games, using a memory size of $10^4$ (on the left) and $10^5$ (on the right). Total rewards calculated after $5\times10^6$ steps, and $3.5\times10^6$ for NEC when using a memory size of $10^5$ per action.}\smallskip
\small
\resizebox{\columnwidth}{!}{%
\begin{tabular}{ c c c  | c c | c c  | c c | c}
\hline
  \textbf{Atari} & \multicolumn{2}{c}{\textbf{MFEC $10^4$} size} & \multicolumn{2}{c}{\textbf{NEC $10^4$} size}& \multicolumn{2}{c}{\textbf{MFEC $10^5$} size} & \multicolumn{2}{c}{\textbf{NEC $10^5$} size} & \multicolumn{1}{c}{\textbf{D3QN}}\\
 \textbf{Game}  & LRU & DkM & LRU & DkM & LRU & DkM & LRU & DkM & \\
 \hline
 \hline 
 
 Bowling & $51.1\pm2.8$&  $\pmb{53.8 \pm2.8}$ & $23.9 \pm6$ &   $\pmb{48 \pm5}$& $\pmb{62.8\pm1.8}$&  $61 \pm2.9$ & $21.6 \pm6.4$ &   $\pmb{43.8 \pm13.5}$ & $30.5 \pm13.6$\\
 
 Ms. Pac-Man & $2660\pm252$&  $\pmb{4143 \pm1153}$ & $3556\pm783$ &   $\pmb{4264\pm1558}$& $\pmb{5245\pm456}$&  $2862\pm1388$ & $\pmb{5225\pm856}$ &   $4092\pm856$ & $2090\pm275$\\
 
 Pong  & $-5.3\pm1.9$ & $\pmb{0.63 \pm5.3}$& $-2.9\pm6.3$ & $\pmb{-0.35\pm4.3}$& $\pmb{18.6\pm1.4}$ & $15.3 \pm2.8$& $11.1\pm3$ & $\pmb{15.9\pm1.6}$ & $6.1\pm9.6$\\
 
 Space Invaders  & $\pmb{472\pm28}$& $408\pm60$& $\pmb{694\pm105}$ & $549\pm42$& $\pmb{806\pm88}$& $430\pm61$& $\pmb{854\pm67}$ & $526\pm88$ & $767\pm72$\\
 
 Q*bert &$1724\pm846$&  $\pmb{1728\pm942}$& $\pmb{2168\pm685}$ & $1556\pm342$&$\pmb{4930\pm787}$&  $1140\pm241$& $\pmb{5093\pm1480}$ & $3389\pm745$&  $4463\pm460$\\
 \hline
\end{tabular}
}
\label{atari_table}
\end{table}

\section{Discussion}
We investigated various methods to improve the memory complexity of EC methods, and proposed a novel online clustering algorithm (DkM) that outperforms other methods for small memory sizes. In simple environments, it can require 10 to 20 times less memory to effectively train the agent, as compared to the original LRU strategy (Tables \ref{table1}, \ref{table2}). We speculate that DkM outperforms the alternative methods because of its ability to reduce redundant information through clustering, and concurrently the ability to adapt to changes in the state distributions, as investigated in Section \ref{sec:embedding}. An interesting observation is that while LRU tends to improve monotonically with an increase in the buffer size (Tables \ref{table1}, \ref{table2}), the online clustering methods do not share that property---investigating this could be an interesting avenue for further understanding the role of clustering representations in EC.

In Atari games, the improvements hold for smaller buffer sizes ($10^4$), but not for larger buffer sizes ($10^5$), where the original LRU strategy performs best. The operation of merging states into clusters could increase state aliasing, making it more difficult for an agent to properly differentiate such situations (e.g., small pixel differences). Possible improvements could be to make the clustering process itself dependent on the $Q$-values.

Overall, DkM allows the use of smaller memory sizes for EC methods, making them more memory- and computationally-efficient. Combined with the sample-efficiency of EC methods, we believe that it is a promising technique to apply to real-world scenarios where resources can be limited and the acquisition of new data expensive.

\bibliography{AAAI}

\begin{thebibliography}{22}
\providecommand{\natexlab}[1]{#1}
\providecommand{\url}[1]{\texttt{#1}}
\expandafter\ifx\csname urlstyle\endcsname\relax
  \providecommand{\doi}[1]{doi: #1}\else
  \providecommand{\doi}{doi: \begingroup \urlstyle{rm}\Url}\fi

\bibitem[Abraham and Robins(2005)]{abraham2005memory}
Wickliffe~C Abraham and Anthony Robins.
\newblock Memory retention--the synaptic stability versus plasticity dilemma.
\newblock \emph{Trends in Neurosciences}, 28\penalty0 (2):\penalty0 73--78,
  2005.

\bibitem[Azzalini and Capitanio(1999)]{azzalini1999statistical}
Adelchi Azzalini and Antonella Capitanio.
\newblock Statistical applications of the multivariate skew normal
  distribution.
\newblock \emph{Journal of the Royal Statistical Society: Series B (Statistical
  Methodology)}, 61\penalty0 (3):\penalty0 579--602, 1999.

\bibitem[Bellemare et~al.(2013)Bellemare, Naddaf, Veness, and
  Bowling]{bellemare2013arcade}
Marc~G Bellemare, Yavar Naddaf, Joel Veness, and Michael Bowling.
\newblock The arcade learning environment: An evaluation platform for general
  agents.
\newblock \emph{Journal of Artificial Intelligence Research}, 47:\penalty0
  253--279, 2013.

\bibitem[Beygelzimer et~al.(2006)Beygelzimer, Kakade, and
  Langford]{beygelzimer2006cover}
Alina Beygelzimer, Sham Kakade, and John Langford.
\newblock Cover trees for nearest neighbor.
\newblock In \emph{{International Conference on Machine Learning}}, pages
  97--104. ACM, 2006.

\bibitem[Blundell et~al.(2016)Blundell, Uria, Pritzel, Li, Ruderman, Leibo,
  Rae, Wierstra, and Hassabis]{blundell2016model}
Charles Blundell, Benigno Uria, Alexander Pritzel, Yazhe Li, Avraham Ruderman,
  Joel~Z Leibo, Jack Rae, Daan Wierstra, and Demis Hassabis.
\newblock Model-free episodic control.
\newblock \emph{arXiv preprint arXiv:1606.04460}, 2016.

\bibitem[Brockman et~al.(2016)Brockman, Cheung, Pettersson, Schneider,
  Schulman, Tang, and Zaremba]{brockman2016openai}
Greg Brockman, Vicki Cheung, Ludwig Pettersson, Jonas Schneider, John Schulman,
  Jie Tang, and Wojciech Zaremba.
\newblock {OpenAI Gym}.
\newblock \emph{arXiv preprint arXiv:1606.01540}, 2016.

\bibitem[Hasselt(2010)]{hasselt2010double}
Hado~V Hasselt.
\newblock {Double Q-learning}.
\newblock In \emph{Advances in Neural Information Processing Systems}, pages
  2613--2621, 2010.

\bibitem[Hayes et~al.(2019)Hayes, Cahill, and Kanan]{hayes2019memory}
Tyler~L Hayes, Nathan~D Cahill, and Christopher Kanan.
\newblock Memory efficient experience replay for streaming learning.
\newblock In \emph{{International Conference on Robotics and Automation}},
  pages 9769--9776, 2019.

\bibitem[Isele and Cosgun(2018)]{isele2018selective}
David Isele and Akansel Cosgun.
\newblock Selective experience replay for lifelong learning.
\newblock In \emph{AAAI Conference on Artificial Intelligence}, 2018.

\bibitem[Johnson and Lindenstrauss(1984)]{johnson1984extensions}
William~B Johnson and Joram Lindenstrauss.
\newblock {Extensions of Lipschitz mappings into a Hilbert space}.
\newblock \emph{{Contemporary Mathematics}}, 26\penalty0 (189-206):\penalty0 1,
  1984.

\bibitem[King(2012)]{king2012online}
Angie King.
\newblock Online k-means clustering of nonstationary data.
\newblock \emph{Prediction Project Report}, pages 1--9, 2012.

\bibitem[Kingma and Welling(2014)]{kingma2013auto}
Diederik~P Kingma and Max Welling.
\newblock {Auto-encoding variational Bayes}.
\newblock In \emph{International Conference on Learning Representations}, 2014.

\bibitem[Lake et~al.(2017)Lake, Ullman, Tenenbaum, and
  Gershman]{lake2017building}
Brenden~M Lake, Tomer~D Ullman, Joshua~B Tenenbaum, and Samuel~J Gershman.
\newblock Building machines that learn and think like people.
\newblock \emph{Behavioral and Brain Sciences}, 40, 2017.

\bibitem[Liberty et~al.(2016)Liberty, Sriharsha, and
  Sviridenko]{liberty2016algorithm}
Edo Liberty, Ram Sriharsha, and Maxim Sviridenko.
\newblock An algorithm for online k-means clustering.
\newblock In \emph{Meeting on Algorithm Engineering and Experiments}, pages
  81--89. SIAM, 2016.

\bibitem[Lin(1992)]{lin1992self}
Long-Ji Lin.
\newblock Self-improving reactive agents based on reinforcement learning,
  planning and teaching.
\newblock \emph{Machine Learning}, 8\penalty0 (3-4):\penalty0 293--321, 1992.

\bibitem[Machado et~al.(2017)Machado, Bellemare, and
  Bowling]{machado2017laplacian}
Marios~C Machado, Marc~G Bellemare, and Michael Bowling.
\newblock {A Laplacian framework for option discovery in reinforcement
  learning}.
\newblock In \emph{International Conference on Machine Learning}, pages
  2295--2304, 2017.

\bibitem[McClelland et~al.(1995)McClelland, McNaughton, and
  O'Reilly]{mcclelland1995there}
James~L McClelland, Bruce~L McNaughton, and Randall~C O'Reilly.
\newblock Why there are complementary learning systems in the hippocampus and
  neocortex: insights from the successes and failures of connectionist models
  of learning and memory.
\newblock \emph{{Psychological Review}}, 102\penalty0 (3):\penalty0 419, 1995.

\bibitem[Mnih et~al.(2015)Mnih, Kavukcuoglu, Silver, Rusu, Veness, Bellemare,
  Graves, Riedmiller, Fidjeland, Ostrovski, et~al.]{mnih2015human}
Volodymyr Mnih, Koray Kavukcuoglu, David Silver, Andrei~A Rusu, Joel Veness,
  Marc~G Bellemare, Alex Graves, Martin Riedmiller, Andreas~K Fidjeland, Georg
  Ostrovski, et~al.
\newblock Human-level control through deep reinforcement learning.
\newblock \emph{Nature}, 518\penalty0 (7540):\penalty0 529, 2015.

\bibitem[Pritzel et~al.(2017)Pritzel, Uria, Srinivasan, Badia, Vinyals,
  Hassabis, Wierstra, and Blundell]{pritzel2017neural}
Alexander Pritzel, Benigno Uria, Sriram Srinivasan, Adria~Puigdomenech Badia,
  Oriol Vinyals, Demis Hassabis, Daan Wierstra, and Charles Blundell.
\newblock Neural episodic control.
\newblock In \emph{{International Conference on Machine Learning}}, pages
  2827--2836, 2017.

\bibitem[Rezende et~al.(2014)Rezende, Mohamed, and
  Wierstra]{rezende2014stochastic}
Danilo~Jimenez Rezende, Shakir Mohamed, and Daan Wierstra.
\newblock Stochastic backpropagation and approximate inference in deep
  generative models.
\newblock In \emph{International Conference on Machine Learning}, pages
  1278--1286, 2014.

\bibitem[Wang et~al.(2016)Wang, Schaul, Hessel, Hasselt, Lanctot, and
  Freitas]{wang2016dueling}
Ziyu Wang, Tom Schaul, Matteo Hessel, Hado Hasselt, Marc Lanctot, and Nando
  Freitas.
\newblock Dueling network architectures for deep reinforcement learning.
\newblock In \emph{International Conference on Machine Learning}, pages
  1995--2003, 2016.

\bibitem[Zhong(2005)]{zhong2005efficient}
Shi Zhong.
\newblock Efficient online spherical k-means clustering.
\newblock In \emph{IEEE International Joint Conference on Neural Networks},
  volume~5, pages 3180--3185. IEEE, 2005.

\end{thebibliography}

\newpage

{\huge{\textbf{Appendix}}}

\section{Properties of Memory Storage Strategies}
\label{sec:embedding}

To better understand the behaviour of our DkN algorithm with respect to the other strategies, we created a simple synthetic 2D dataset and and evaluated the $k$-means, online $k$-means, DkN and LRU methods on this data.

For the online methods, we simulated an online data stream first by a series of 2D points uniformly distributed on a grid, followed by a series of 2D points from a skew normal distribution \cite{azzalini1999statistical}, which represents random exploration of the state space, and then convergence. In order to qualitatively explore the properties of the different methods, we plotted a heatmap of the distribution of states using kernel density estimation.

\begin{figure}[h]
\centering
\begin{subfigure}{0.7\columnwidth}
  \includegraphics[width=1\textwidth]{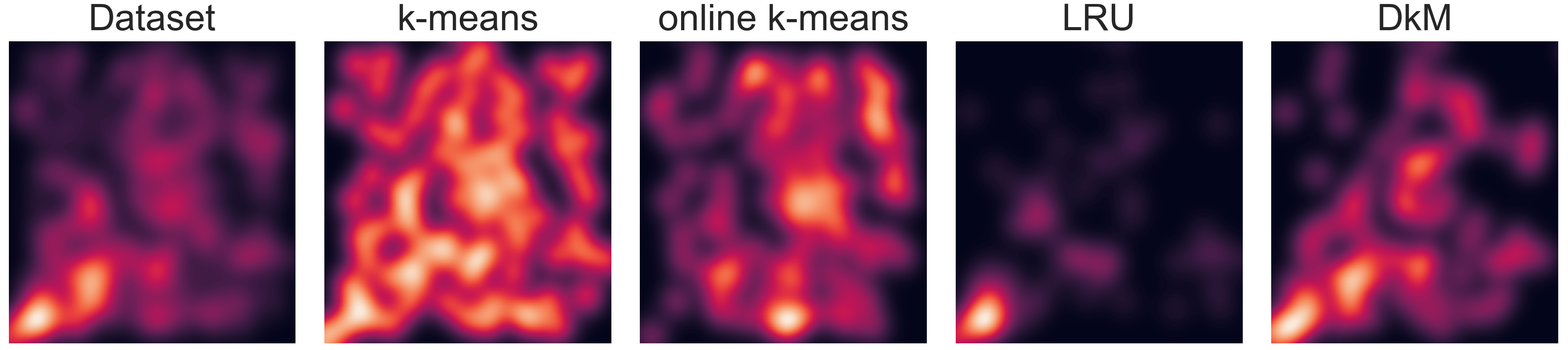} 
  \caption{Comparison between different methodologies}
  \label{fig:kernel_km}
 \end{subfigure}
\begin{subfigure}{0.7\columnwidth}
  \includegraphics[width=1\textwidth]{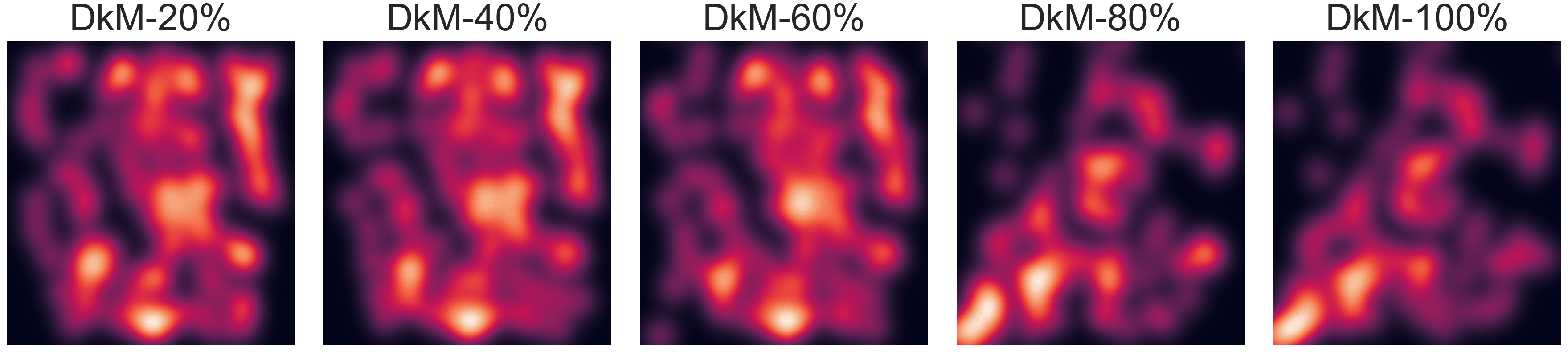} 
  \caption{Evolution of the embedding with DkM during the data stream}
  \label{fig:dkm}
 \end{subfigure}
\caption{Comparison between state distributions generated from the different techniques. DkM shows the best approximation to the real dataset. Kernel density estimates have been fit separately for each method.}
\end{figure}

In Figure \ref{fig:kernel_km} we notice that different strategies cover different parts of the data distribution. $k$-means applied to the whole dataset distributes the clusters across the entire support of the data independently of when the data was first observed. kM and LRU show completely different distributions; the former favours the first data distribution (uniform) due to the static nature of clusters, while the latter favours the second data distribution (skew normal) after forgetting the least-frequently-visited states. In comparison, DkM is able to retain old states, while biasing clusters towards the new distribution. Figure \ref{fig:dkm} shows the evolution of the clusters in DkM after observing $25$, $50$, $75$ and $100\%$ of the dataset. DkM behaves like a mix between kM, where the estimated data distribution covers the support well, and LRU, which prioritises new observations.

\newpage
\section{Classic Control}
\label{sec:classic}

\begin{figure}[H]
\includegraphics[width=1\textwidth]{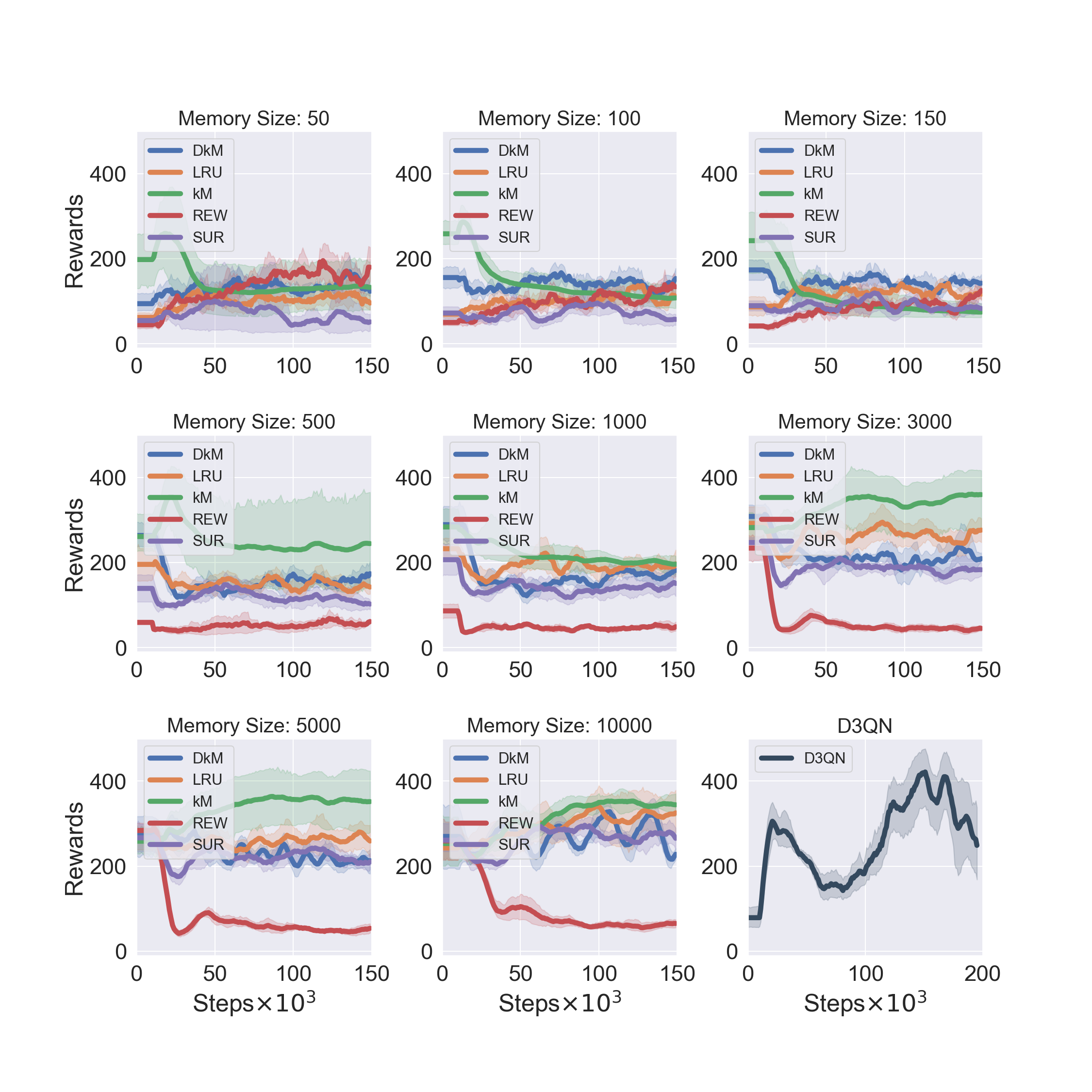}
\centering
\caption{Learning curves of Cartpole for different memory sizes, using MFEC and D3QN}
\label{fig:mfec_cartpole}
\end{figure}

\begin{figure}[H]
\includegraphics[width=1\textwidth]{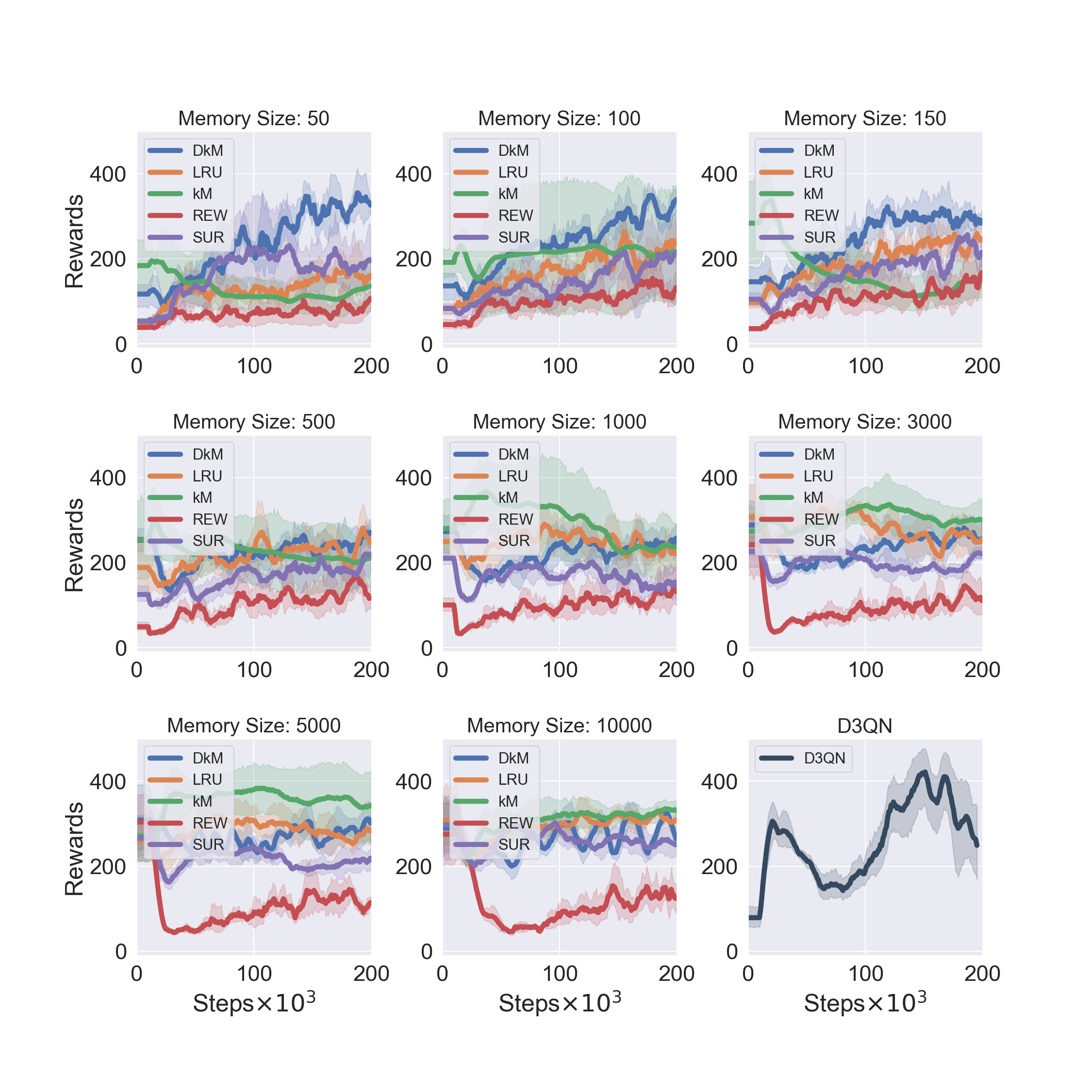}
\centering
\caption{Learning curves of Cartpole for different memory sizes, using NEC and D3QN}
\label{fig:nec_cartpole}
\end{figure}

\begin{figure}[H]
\includegraphics[width=1\textwidth]{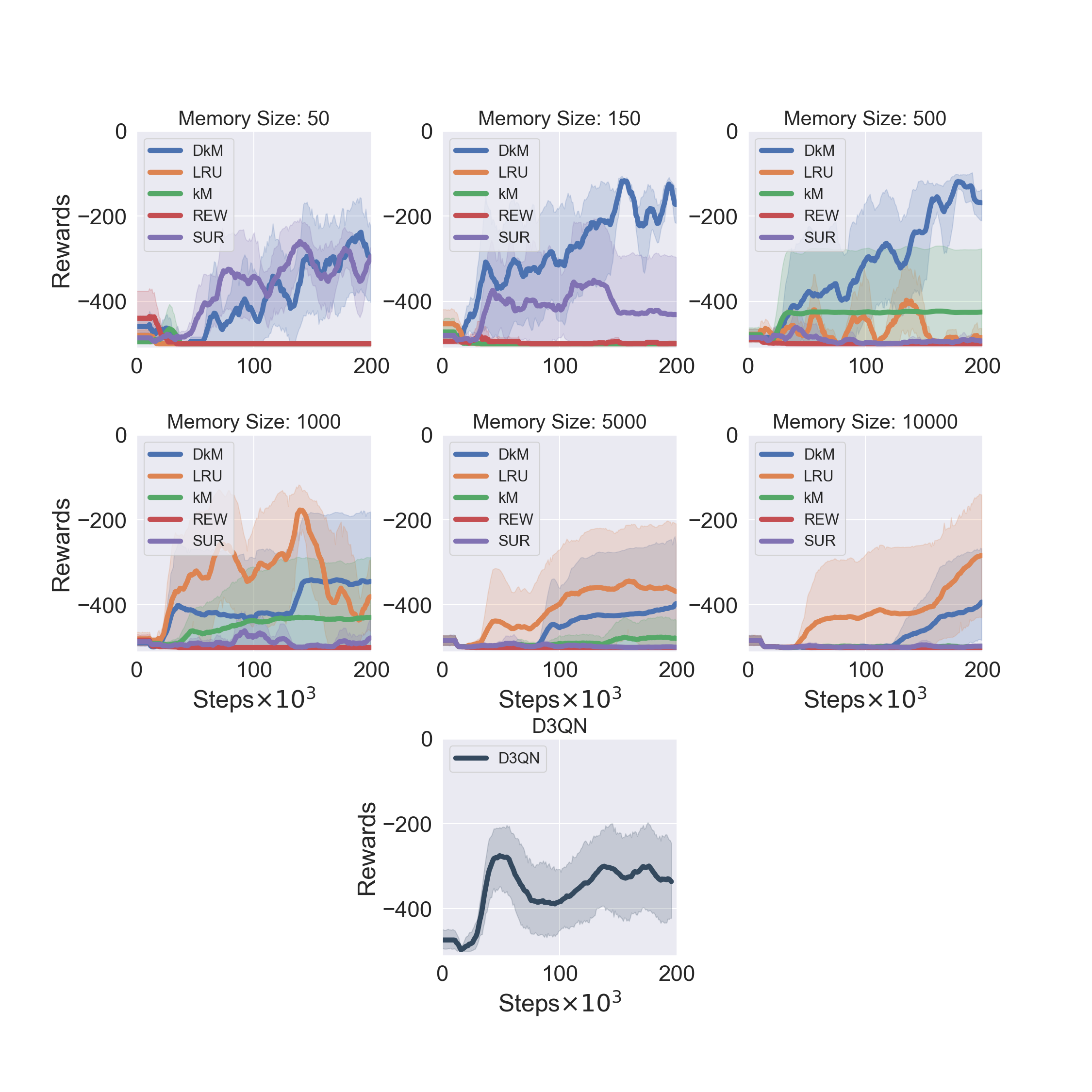}
\centering
\caption{Learning curves of Acrobot for different memory sizes, using MFEC and D3QN}
\label{fig:mfec_acrobot}
\end{figure}

\begin{figure}[H]
\includegraphics[width=1\textwidth]{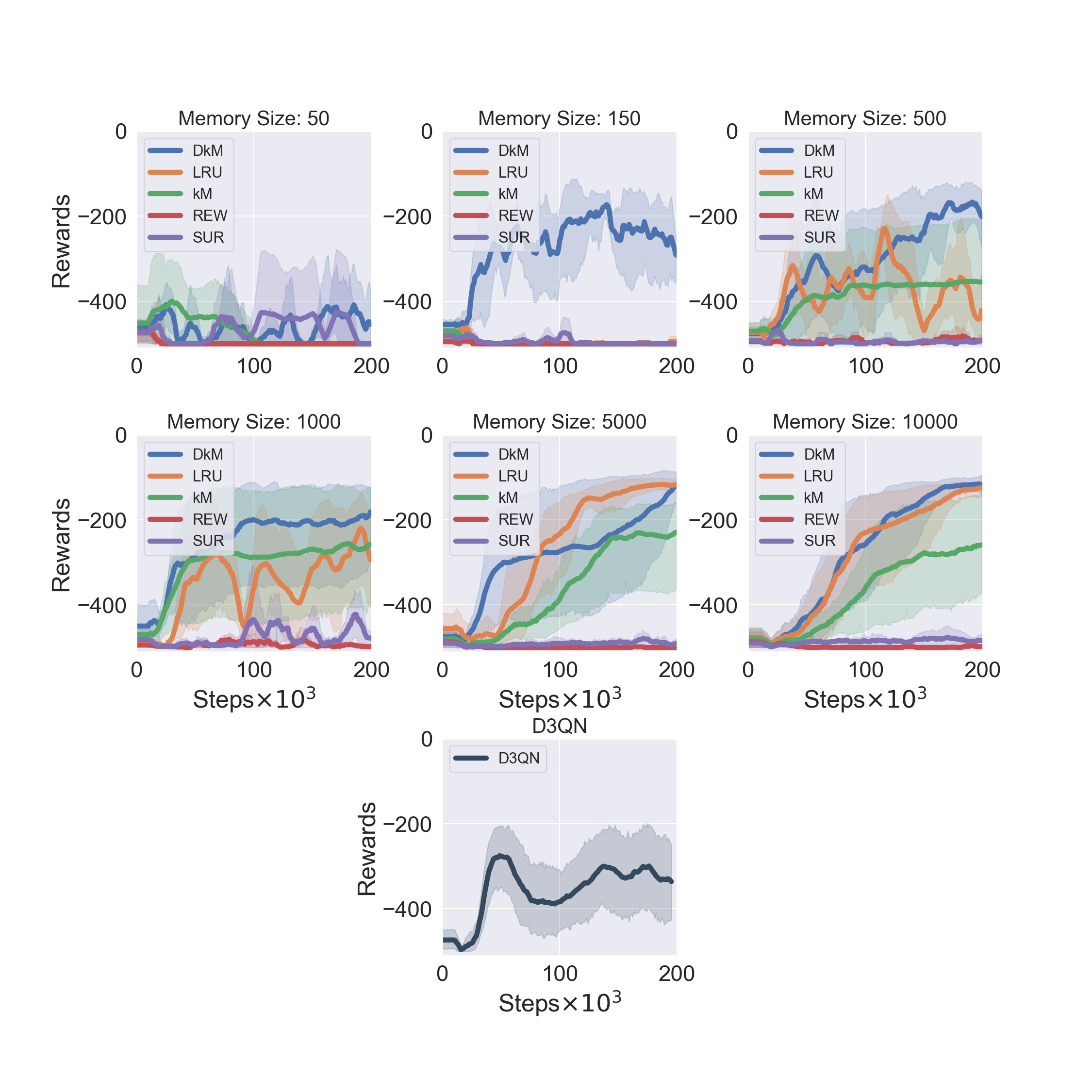}
\centering
\caption{Learning curves of Acrobot for different memory sizes, using NEC and D3QN}
\label{fig:nec_acrobot}
\end{figure}

\newpage
\section{Gridworld}
\label{sec:grid}

\begin{figure}[H]
\includegraphics[width=1\textwidth]{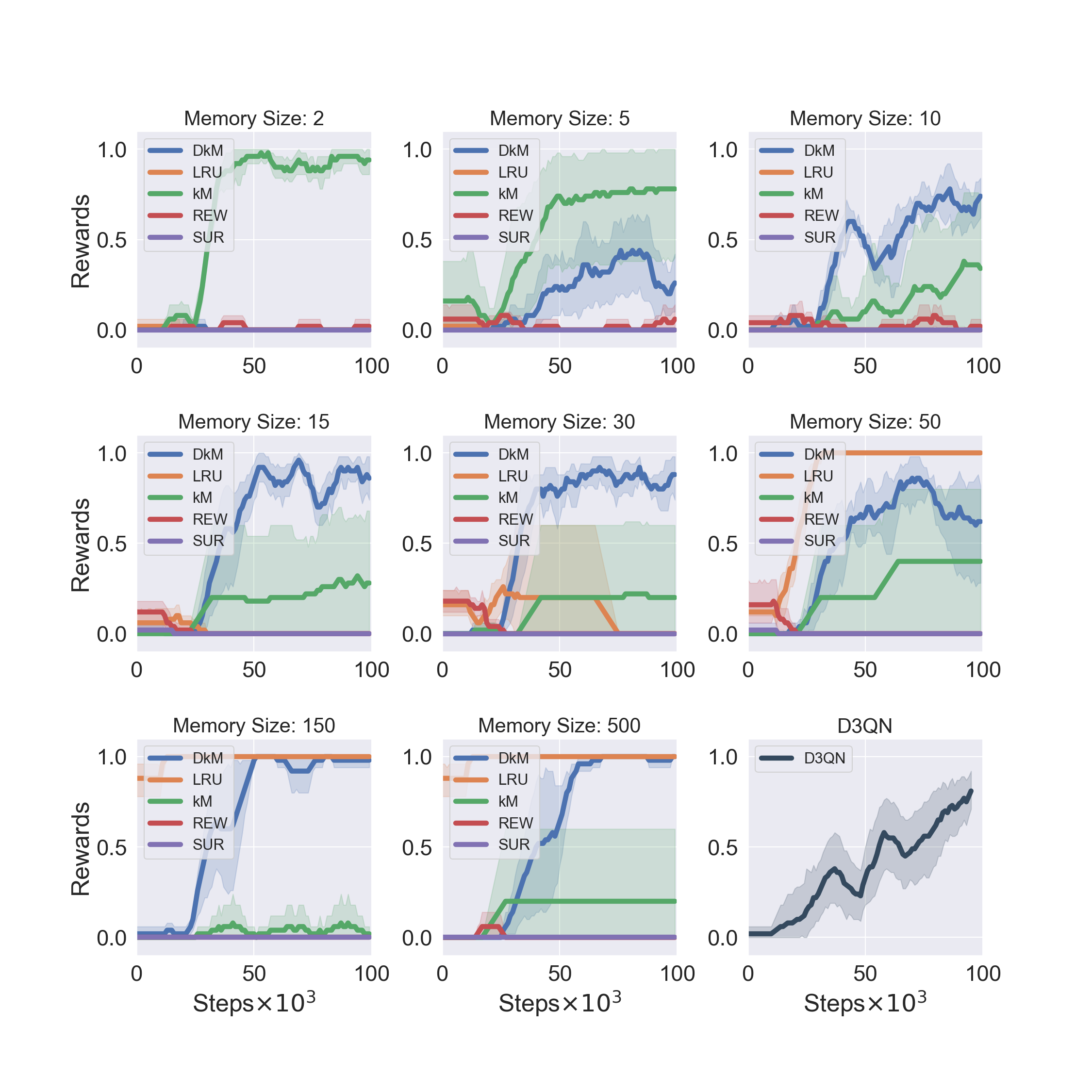}
\centering
\caption{Learning curves of OpenRoom for different memory sizes, using MFEC and D3QN}
\label{fig:mfec_openroom}
\end{figure}

\begin{figure}[H]
\includegraphics[width=1\textwidth]{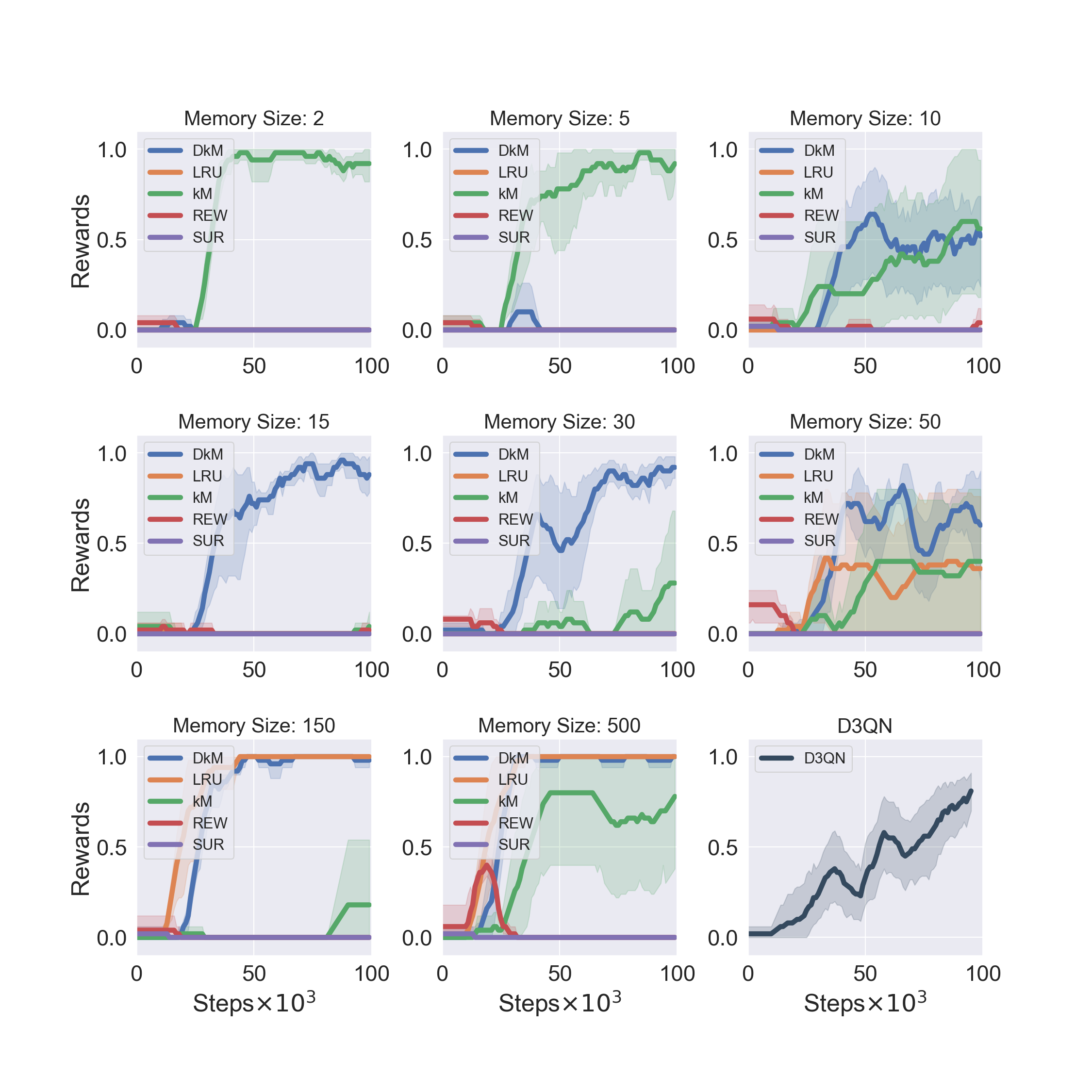}
\centering
\caption{Learning curves of OpenRoom for different memory sizes, using NEC and D3QN}
\label{fig:nec_openroom}
\end{figure}

\begin{figure}[H]
\includegraphics[width=1\textwidth]{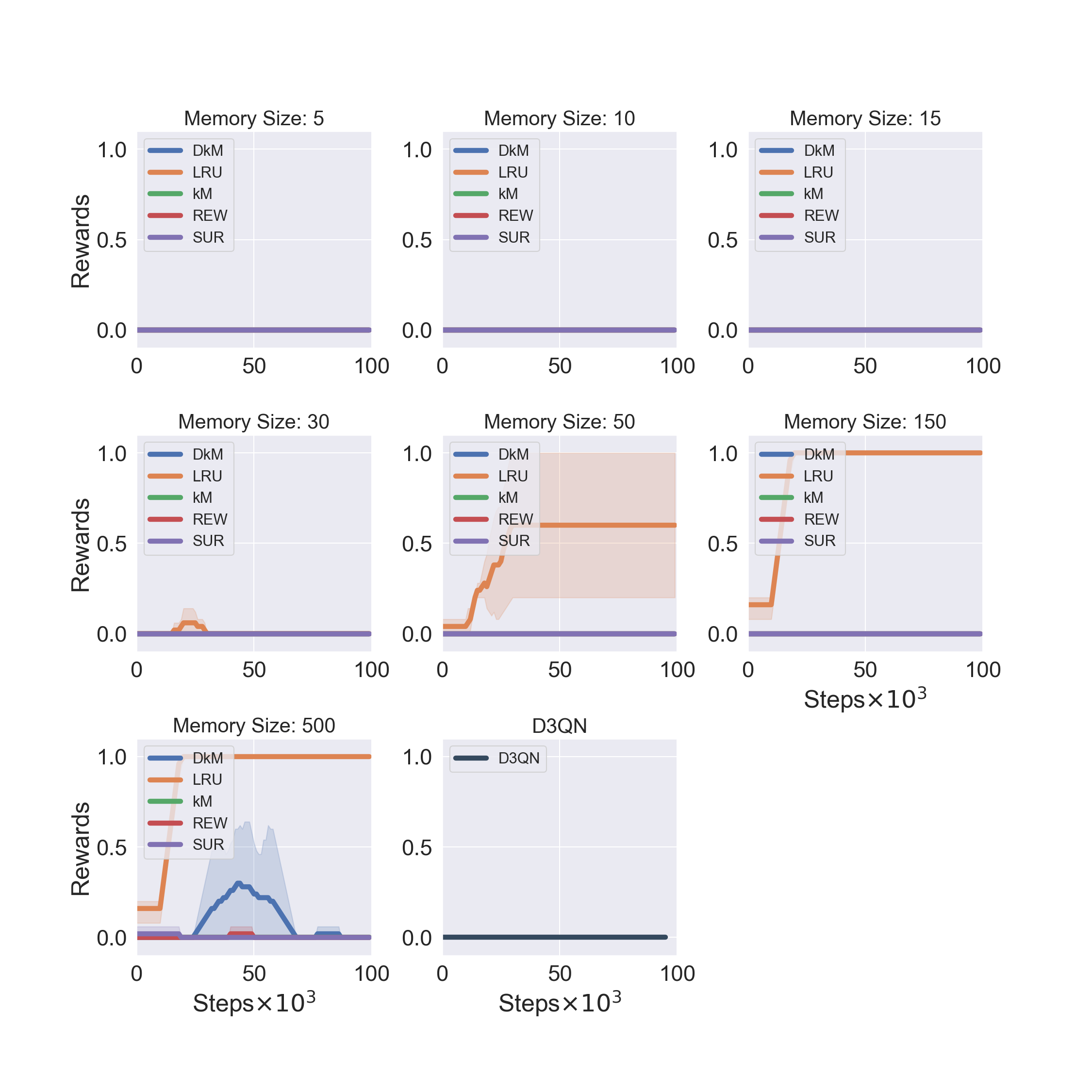}
\centering
\caption{Learning curves of FourRoom for different memory sizes, using MFEC and D3QN}
\label{fig:mfec_fourroom}
\end{figure}

\begin{figure}[H]
\includegraphics[width=1\textwidth]{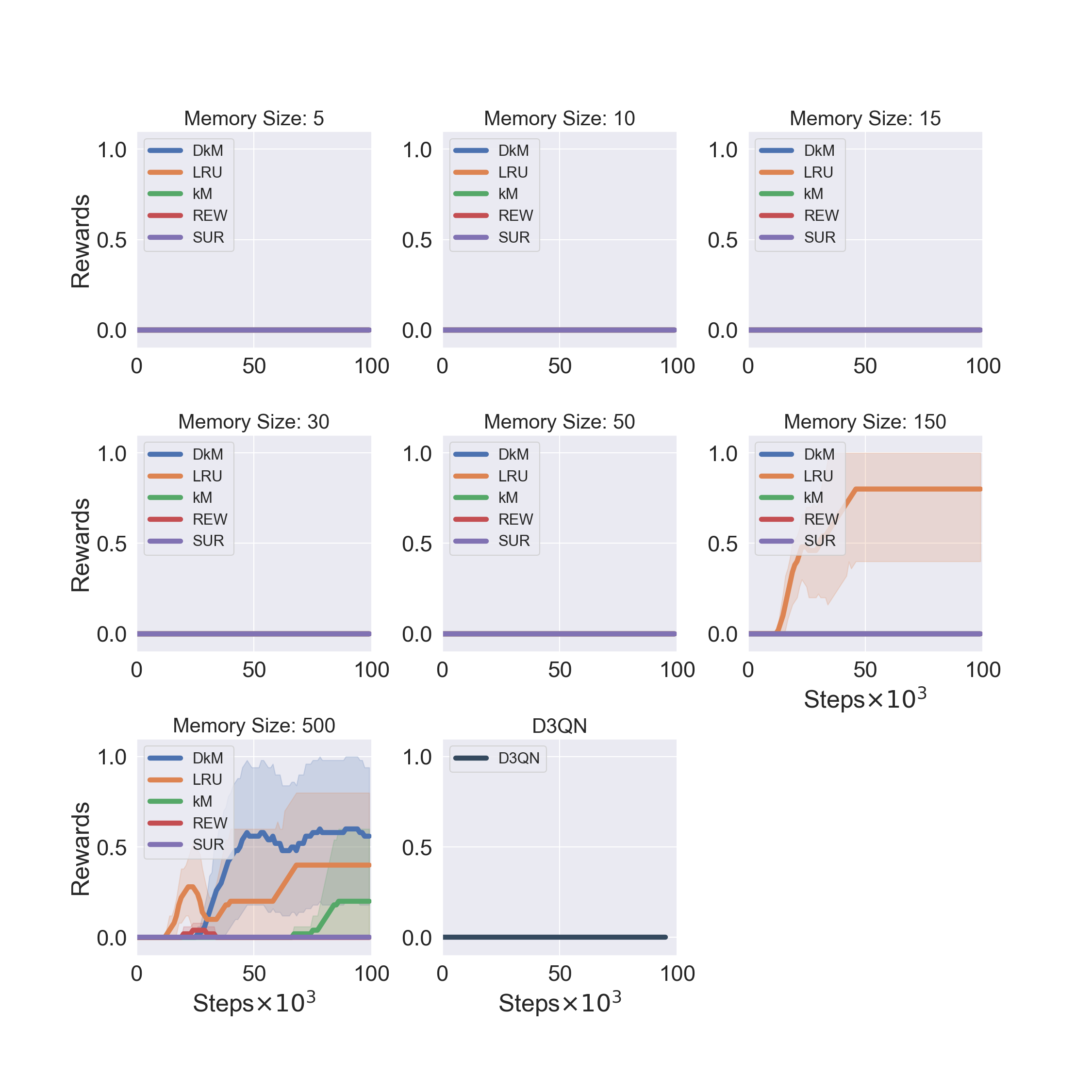}
\centering
\caption{Learning curves of FourRoom for different memory sizes, using NEC and D3QN}
\label{fig:nec_fourroom}
\end{figure}

\newpage
\section{Atari games}
\label{sec:atari_app}

\begin{figure}[H]
\centering
  \includegraphics[width=0.6\textwidth]{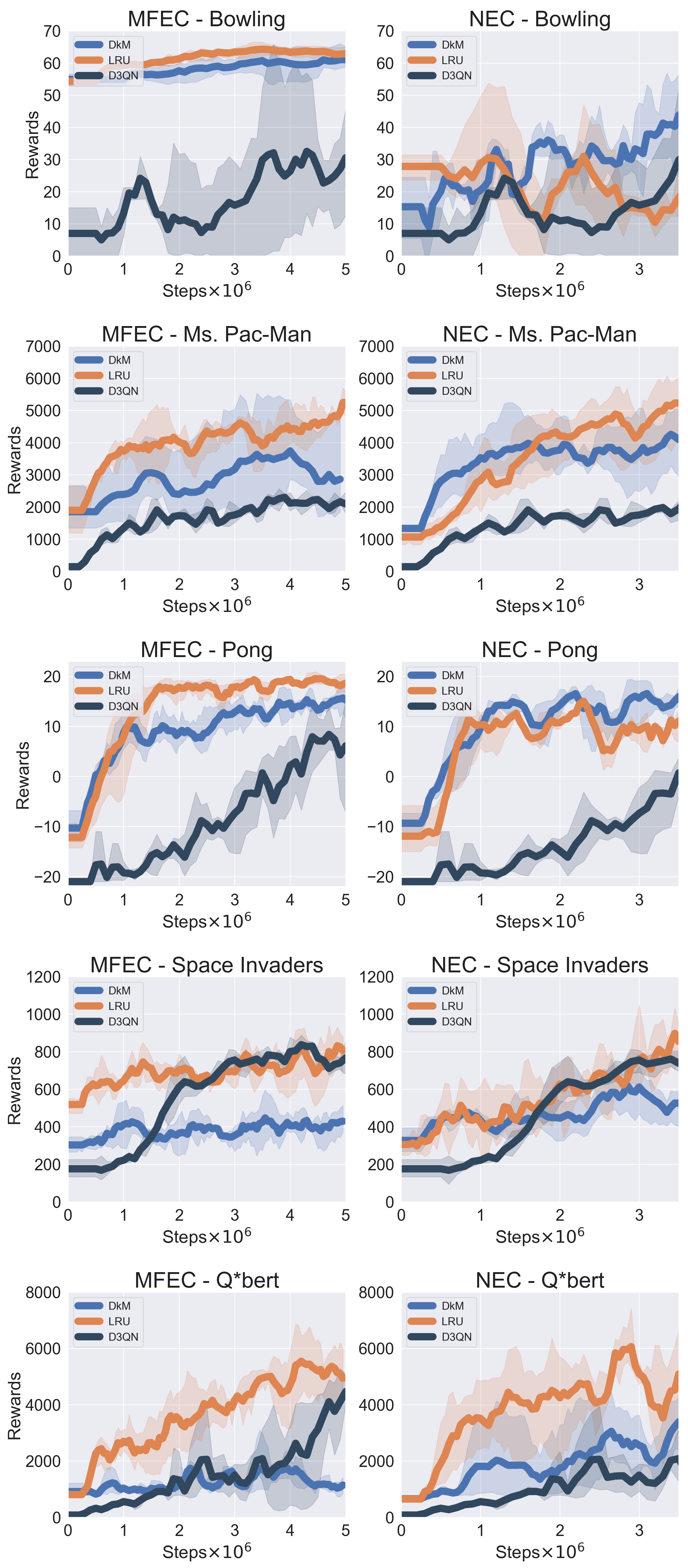} 
   \label{atari100}
    \caption{Learning curves of 5 Atari games using MFEC, NEC and D3QN. EC methods with a memory size of $10^5$ per action.}
\end{figure}

\newpage
\section{Hyperparameters}
\label{sec:hyperparameters}

\begin{table}[H]
\centering
\small
\caption{Hyperparameters used with MFEC across all the environments.}\smallskip
\begin{tabular}{ c  c  c c  }
\hline
  \textbf{Parameters name} & \multicolumn{1}{c}{\textbf{Classic Control}} & \multicolumn{1}{c}{\textbf{Room Domains}} & \multicolumn{1}{c}{\textbf{Atari Games}}\\
 \hline
 \hline
 Number of neighbours $k$&$11$&$11$&$11$\\ 
  
 $\epsilon$ initial &$1$&$1$&$1$\\
 
 $\epsilon$ final&$5\times10^{-3}$&$5\times10^{-3}$&$5\times10^{-3}$\\
 
$\epsilon$ anneal start (steps) &$5\times10^3$&$5\times10^3$&$5\times10^3$\\
 
$\epsilon$ anneal end (steps)&$25\times10^3$&$25\times10^3$&$25\times10^3$\\
 
 Discount factor $\lambda$&$0.99$&$0.99$&$0.99$\\
 
 Reward clip &None&None&None\\
  
Kernel delta $\delta$&$10^{-3}$&$10^{-3}$&$10^{-3}$\\

Observation projection & None& None& Gaussian\\

Projection key size & None& None& $128$\\

 \hline
\end{tabular}
\label{par_MFEC}
\end{table}

\begin{table}[H]
\centering
\small
\caption{Hyperparameters used with NEC across all the environments.}\smallskip
\begin{tabular}{ c  c  c  c  }
\hline
  \textbf{Parameters name} & \multicolumn{1}{c}{\textbf{Classic Control}} & \multicolumn{1}{c}{\textbf{Room Domains}} & \multicolumn{1}{c}{\textbf{Atari Games}}\\
 \hline
 \hline
 Number of neighbours $k$&11&11&50\\
 
 Experience replay size &$10^5$&$10^5$&$10^5$\\
 
Memory learning rate $\alpha$ &$0.1$&$0.1$&$0.1$\\

RMSprop learning rate &$7.92\times10^{-6}$&$7.92\times10^{-6}$&$7.92\times10^{-6}$\\
  
RMSprop momentum&$0.95$&$0.95$&$0.95$\\

RMSprop $\epsilon$ &$10^{-2}$&$10^{-2}$&$10^{-2}$\\
 
 $\epsilon$ initial&$1$&$1$&$1$ \\
 
 $\epsilon$ final &$5\times10^{-3}$&$5\times10^{-3}$&$10^{-3}$\\
 
$\epsilon$ anneal start (steps) &$5\times10^3$&$5\times10^3$&$5\times10^3$\\
 
$\epsilon$ anneal end (steps) &$25\times10^3$&$25\times10^3$&$25\times10^3$\\
 
 Discount factor $\lambda$ &$0.99$&$0.99$&$0.99$\\
 
 Reward clip &None&None&None\\
  
Kernel delta $\delta$&$10^{-3}$&$10^{-3}$&$10^{-3}$\\

Batch size&$32$&$32$&$32$\\

$n$-step return&$100$&$100$&$100$\\

Key size &$64$&$64$&$128$\\

Training start (steps)&$10^3$&$10^3$&$5\times10^4$\\
 
 \hline
\end{tabular}

\label{par_NEC}
\end{table}

\begin{table}[H]
\centering
\small
\caption{Hyperparameters used with D3QN across all the environments.}\smallskip
\begin{tabular}{ c  c  c  c  }
\hline
  \textbf{Parameters name} & \multicolumn{1}{c}{\textbf{Classic Control}} & \multicolumn{1}{c}{\textbf{Room Domains}} & \multicolumn{1}{c}{\textbf{Atari Games}}\\
 \hline
 \hline
 
 Experience replay size &$10^5$&$10^5$&$10^6$\\
 
 RMSprop learning rate &$25\times10^{-5}$&$25\times10^{-5}$&$25\times10^{-5}$\\
RMSprop momentum &$0.95$&$0.95$&$0.95$\\

RMSprop $\epsilon$ &$10^{-2}$&$10^{-2}$&$10^{-2}$\\
 
 $\epsilon$ initial &$1$&$1$&$1$ \\
 
 $\epsilon$ final &$5\times10^{-3}$&$5\times10^{-3}$&$10^{-2}$\\
 
$\epsilon$ anneal start (steps)  &$1$ &$1$ &$1$\\
 
$\epsilon$ anneal end (steps) &$5\times10^4$&$5\times10^4$&$10^6$\\
 
 Discount factor $\lambda$&$0.99$&$0.99$&$0.99$ \\
 
 Reward clip &Yes&Yes&Yes\\
  
Batch size&$32$&$32$&$32$\\

Training start (steps) &$5\times10^3$&$5\times10^3$&$12.5\times10^3$\\

 Target network update (steps) &$7.5\times10^3$&$10^3$&$10^3$\\

 \hline
\end{tabular}
\label{par_DQN}
\end{table}

\end{document}